# Reddit is all you need: Authorship profiling for Romanian

ECATERINA ŞTEFĂNESCU and ALEXANDRU-IULIUS JERPELEA

*"Tudor Vianu" National College of Computer Science, Bucharest, Romania*
*ecaterina@stefanescu.eu, alex.jerpelea@gmail.com*

## Abstract

Authorship profiling is the process of identifying an author's characteristics based on their writings. This centuries old problem has become more intriguing especially with recent developments in Natural Language Processing (NLP). In this paper, we introduce a corpus of short texts in the Romanian language, annotated with certain author characteristic keywords; to our knowledge, the first of its kind. In order to do this, we exploit a social media platform called Reddit. We leverage its thematic community-based structure (subreddits structure), which offers information about the author's background. We infer an user's demographic and some broad personal traits, such as age category, employment status, interests, and social orientation based on the subreddit and other cues. We thus obtain a 23k+ samples corpus, extracted from 100+ Romanian subreddits. We analyse our dataset, and finally, we fine-tune and evaluate Large Language Models (LLMs) to prove baselines capabilities for authorship profiling using the corpus, indicating the need for further research in the field. We publicly release all our resources.[1]

## 1 Introduction

Author-specific traits such as demographic characteristics (age, gender, educational background, etc) (Akçay, 2017), personality traits (Ahmed and Naqvi., 2015), interests, and disciplinary background (Coffin and Hewings, 2003) all influence the way humans communicate and reflect in the language used. This includes written text, which this paper aims to study.

Authorship profiling is one of the three categories within the broader field of authorship analysis (the other two being authorship attribution and verification) (Misini et al., 2022) with the purpose of building the profile of a text's author (Lundeqvist and Svensson, 2017). This can be done by analysing the information extracted from a text through the following approaches (Pervaz et al., 2015; Reddy et al., 2016): (1) stylometric (Grivas et al., 2015), which analyses writing style, (2) content-based, (3) topic-based, and (4) mixed. The characteristics that one may want to identify about the author can vary, depending on the nature of the task. They can range from demographic characteristics (Reddy et al., 2016), such as age, gender, and native language, to psychographic characteristics (Pervaz et al., 2015), such as extraversion or neuroticism. We deploy a mixed approach that aims to identify as many keywords from the described categories (if possible), relying on Large Language Models (LLMs) to extract them. We aim that the LLM learns this behaviour from the very nature of our collected dataset.

Social media has become an essential medium of modern communication, enabling people to share opinions and content within specific contexts (Schrape, 2016). In today's world, it has become a vital platform for human interaction, making it easier for people with common interests and traits to connect. Among these spaces, Reddit is one of the best at representing these aspects, as users can share interests and engage with like-minded individuals (Green and McCann 24-28). This is possible due to the platform's structure, functioning on the basis of subreddits, self-formed communities of users brought together by a certain topic, interest, and, most importantly, certain character traits (Medvedev et al., 2019; Green and McCann 24-28). By manually constructing a set of general characteristics that define a certain subreddit's users, we are able to annotate posts with certain keywords that relate to that user, otherwise named as the author.

---

[1] https://huggingface.co/roship-profiling, Accessed: 2024-10-12

Through this paper, we aim to advance the capabilities of authorship profiling technology for the Romanian language. To our knowledge, no such systems have been specifically designed for the Romanian language.

More than that, we note the scarcity of valuable data in this regard for the Romanian language. We have addressed this issue by building our own dataset, based on Reddit posts within subreddits, using their corresponding topics to infer useful information about the authors. We have decided to analyse the text from the perspective of the following categories: (1) subdialect of the Daco-Romanian dialect, (2) educational or employment status, (3) personal interests, such as computer science and music, and (4) personal inclinations, such as social orientation and technological interest.

We then finetune a LLM for the task. However, given that the attribution of a clear attribute for each category is not always possible, the LLM can be rather viewed as an author-related keyword generator.

Our work to fill the existing gap can thus be summarised as follows:

- A corpus of Reddit posts, comments, and other texts in the Romanian language, annotated with keywords, that correspond to certain categories, and which relate to their author's background/characteristics.
- A fine-tuned Llama 3.1-8B Instruct model (Dubey et. al., 2024) that generates such keywords from text.

In this paper, we first describe the data collection/annotation process and our baseline LLM development. In the end, we discuss the limitations of our low-resource background and the need for further research in the emerging field of Romanian authorship profiling.

## 2 Related Work

With recent technological developments, authorship profiling has evolved in the past years (Chinea-Rios et al., 2022), reaching and developing for languages different from English. There are various authorship profiling applications for a diverse set of languages, including Arabic (Bassem et al., 2024), Spanish (Villegas et al., 2014; García-Díaz et al., 2024), Lithuanian (Kapočiūtė-Dzikienė et al., 2015), and Russian (Litvinova et al., 2016.). In most cases, the tasks involve the determination of age and gender (HaCohen-Kerner, 2022) or other demographic traits (Reddy et al., 2016). Usually, if the systems target personality traits (Ahmed and Naqvi., 2015), they are usually regarding the five mainly researched ones, namely extroverted, stable, agreeable, conscientious, and open (Iqbal et al., 2015). Research regarding personal inclinations is not as common. A serious issue impacting all approaches is the lack of available data on the correlation between an author's characteristics and his written text. This causes even more concern for less-resourced languages.

To our knowledge, there are not any approaches that leverage Reddit for an authorship profiling related corpus construction. One of the most notable authorship analysis attempts focuses on the use of other social media platforms related to authorship attribution (Casimiro et al., 2020; Matias and Digiampietri, 2023).

A significant work involving Romanian and Reddit is the first dataset targeting the popularity prediction of Romanian Reddit posts (Rogoz et al., 2024). Our corpuses' building steps share similarities, but our approaches differ. For example, when scrapping Reddit, we only used Reddit's API. This API has a limit of 1000 posts per subreddit. Rogoz et al. (2024) have overcome this limitation by using an open-source archive. The numbers of data samples of the dataset and our corpus are very close (28107 and 26517 respectively). However, the ratio between the number of subreddits considered and the number of posts extracted per subreddit is very different. Rogoz et al. 's (2024) corpus only targets 5 Romanian subreddits, while we tackle 146.

Another noteworthy development in authorship analysis involving the Romanian language involves authorship profiling's more popular counterpart, authorship attribution (Nitu and Dascalu, 2024).

Lastly, Ciobanu and Dinu (2016) have tackled automatic dialect identification for Romanian. Our LLM, besides other author characteristics, does a similar task. However, we only identify varieties of Daco-Romanian and do not approach the broader Romanian dialectal continuum.

## 3 Methodology

### 3.1 Finding Romanian Subreddits

Due to the nature of tag attribution to Reddit posts that belong to a certain subreddit, establishing a list of valid subreddits is an important step. After the scouting process, we have identified a list of 263 Romanian subreddits. Among these, 159 were manually distinguished through general searches or sidebar suggestions, while the other 104 were obtained with the help of a Reddit exploration tool that identifies a network of subreddits based on an input subreddit name. A chosen subreddit must be relevant and centred toward the certain topic(s) or traits that most of its authors have, in order to make the tagging process effective. Other criteria were the number and the quality of the posts. Candidates not meeting these requirements were removed. This process results in a total of 146 subreddits.

### 3.2 Characterising the Subreddits

#### 3.2.1 Subdialects

This tag was given to subreddits corresponding to Romanian cities and regions, where upon a manual analysis, speakers are likely to write in a region-specific register. We have taken into consideration four big varieties of Daco-Romanian: Transilvanian, Moldavian, Oltenian, and Wallachian. As can be seen in Table 2, the distribution is somewhat uniform, except for the Oltenian subdialect which has significantly fewer posts. In Table 1 are the subreddit names that have been assigned to each subdialect (they mostly correspond to cities and regions).

| **Daco-Romanian variety** | **Subreddits** |
|---|---|

| Transilvanian subdialect | albaiulia, Arad, bistrita, cluj, Jobs_Timisoara, mures, Oradea, satumare, TarguMures, timisoara, Timisoara_night_life, Zalau |
|---|---|
| Moldavian subdialect | Bacau, Botosani, iasi, moldova, Moldoveneste, Moldovenii, PiatraNeamt, Suceava |
| Wallachian subdialect | bragadiru, brasov, bucuresti, constanta, pitesti, Ploiesti |
| Oltenian subdialect | Craiova, Oltenia |

**Table 1.** Attribution of regional speech variety to various subreddits

| **Subdialect** | **Number of data samples** |
|---|---|
| Moldavian | 1744 |
| Transilvanian | 2632 |
| Wallachian | 2200 |
| Oltenian | 361 |

**Table 2.** The distribution of data samples targeted by the subdialect tag within its categories

| **Status** | **Number of data samples** |
|---|---|
| Pre-University students | 965 |
| University students | 2670 |
| Employed | 2247 |

**Table 3.** The distribution of data samples targeted by the status tag within its categories

### 3.2.1 Status

This type of tag refers to the author's education or employment status, and because they account for different stages in life, they can be related to some broad age brackets. The characteristics of the subreddits assigned with this type of tag were, for each case, the following:

- Pre-university students - subreddits corresponding to a high school or to the Romanian baccalaureate, the exam to graduate high school,
- University students - subreddits corresponding to certain academic fields, universities, and educational apps,
- Employed - subreddits corresponding to jobs and employment.

In this case, the information regarding the first category is not as prevalent as the other two, due to the nature of the subreddits taken into consideration for analysing, as Table 3 demonstrates.

### 3.2.3 Labels and Field of Interest

As subreddits are commonly centred around a certain concept, the users within that community share an affinity for the topic. Based on almost every subreddit's context, we have identified one or more labels that indicate a certain general interest displayed by the users (authors). We obtain a number of 32 labels that are assigned to each post based on the subreddit that it originates from. Together, these topics form a very diverse set of subjects, but they can also be closely related, like mathematics and physics, or conservatism and liberalism. For such cases, besides the interests of the authors, we have come up with a new category: a *field of interest*. Only 6 in number, their purpose is to group similar labels into broader classes that once again indicate a more general interest of the author. Table 4 showcases this grouping. In case the attribution of specific labels fails, then we aim to eliminate mischaracterization at least with the *field of interest* broader approach.

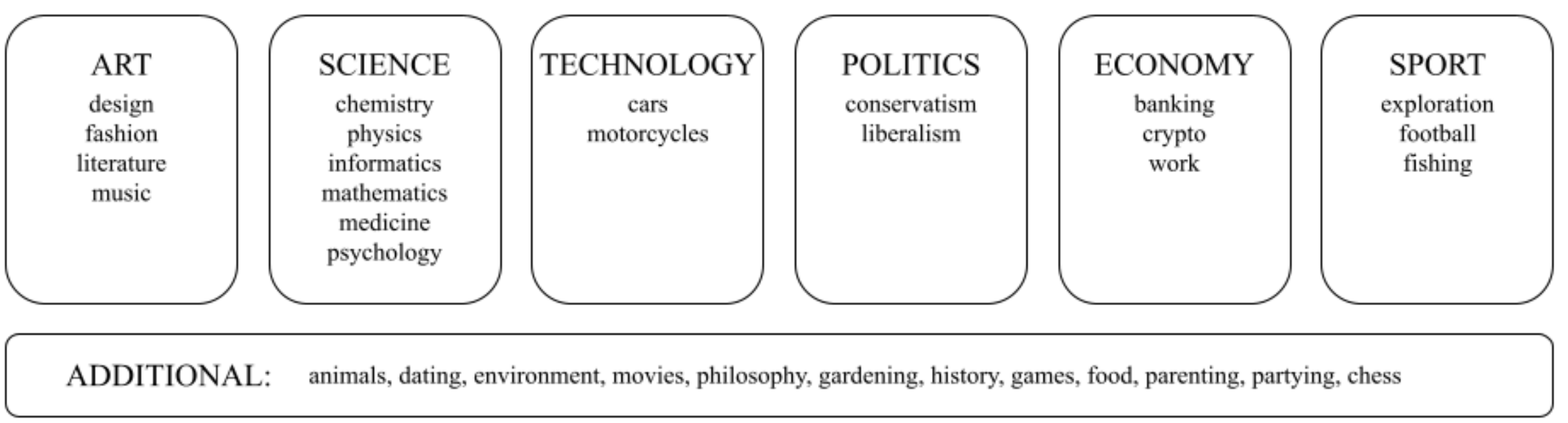

**Table 4.** An illustration showcasing the relations between labels and fields of interest.

### 3.2.4 Personal Inclination

Moreover, with the help of the *field of interest* category, we can construct *personal inclination* tags. These tags aim to encompass every aforementioned label and are adapted to be more suitable for assigning attributes to an author's profile. Multiple such tags can be assigned to a post, based on how its corresponding subreddit has been characterised. This results in 12 types: active lifestyle, arts-oriented, hobbyist, interested in economy, interested in games, interested in health, interested in history or politics, interested in nature, interested in science, interested in technology, media enthusiast, and socially oriented. We have to acknowledge these tags are more related to the text's content, but their correlation to the author's background is important.

## 3.3 Reddit Post Extraction

The posts were gathered using the Reddit API through the Python Reddit API Wrapper (PRAW) library. This method implies a limitation of up to 1000 posts per subreddit or per subreddit category. We aim to keep a balanced dataset anyway so this is not an obstacle. The script checks each post for text content. The information we extracted for each post consists of the title and the body of the post (if it exists). For posts without a body, the title is all that remains. It is added to the dataset only if it has at least 75 characters. We settled on this restraint to ensure that it conveyed a message and that it was related to a certain topic. Once all the relevant information is collected, it is compiled into a CSV file. The resulting dataset has 29,317 annotated posts.

## 3.4 Dataset Cleaning

Posts in the English language are one of the main problems we encountered with the first version of the dataset, compromising our approach aimed at the Romanian language. To eliminate them, we used Google's langdetect library to separate between different languages through two approaches: (1)

identifying Romanian text (25512 rows) and removing the rest (1805 rows), and (2) identifying English text (1256 rows) and eliminating them, resulting in a 28060 rows dataset. Reddit posts are informal in nature, as Romanian written text can oftentimes be mixed with English words or phrases. We chose to be tolerant in this case, and upon manual inspection, the second method yielded better results for our purpose.

Additionally, we have removed excessive whitespaces and compressed multiple consecutive newlines to a maximum of two. Then, we removed URLs, by identifying patterns generally associated with web addresses. If the text of the body consisted only of URLs, we checked the length of the title once again and kept the row in the dataset only if it had a minimum of 75 characters. Stray symbols and rogue parentheses were eliminated as well. After these modifications, the dataset consists of 26,517 data samples. In order to preserve as much linguistic information as possible, we chose to not make any more changes to the dataset.

### 3.5 Final Dataset Construction

After crawling the Reddit posts, note that sometimes, we are not able to generate keywords for each of the presented categories (*subdialect*, *status*, *interest labels*, *field of interest*, *personal inclination*). For example, it is impossible to infer for certain subreddits what regional speech is used there.

Seeking to mitigate this issue and provide a more simplistic structure for the LLM to learn, we merge all keywords from each category into a single set of author-related keywords. The keywords do appear in the same order described in this paper. With this less structured approach, we also shift our authorship profiling task from a classification problem to a feature extraction one.

## 4 Authorship Profiling

Followingly, we use the processed corpus to build a baseline authorship profiling tool that can gather key aspects about a text's author in the form of keywords. As mentioned in Section 3.5, these keywords are unorganised, and are generated only when relevant. For example, if there can not be said anything about the subdialect a writer uses, such keywords should not be generated.

### 4.1 Model Training

We leverage Large Language Models (LLMs), which have proven useful in numerous NLP tasks (Kocmi et al., 2024). Specifically, we seek to utilise their capabilities for feature extraction tasks. The LLM we train is given the task of extracting author-specific keywords.

We experimented with the Llama 3.1-8B-Instruct (Dubey et. al., 2024). It is the most recent LLaMA model, and it covers a diverse set of instructions. Romanian language, although not officially supported, is covered in about 15% of the multilingual tokens from the dataset used in pre-training.

Given its zero or few-shot learning capabilities, we expect the model to perform well even in the conditions of parameter-efficient finetuning. Specifically, we employ LoRA (Low-Rank Adaptation of Large Language Models) fine-tuning (Hu, Edward J. et al., 2021). Instead of updating all model parameters, LoRA focuses on fine-tuning only a small set of parameters through the decomposition of low-rank matrices, which are later added to the model weight matrices.

We are thus able to afford computation, by training around 10% of the number of parameters which would usually be updated during a full fine-tune of the Llama 3.1-8B-Instruct. This has both benefits, as the model will diverge less from its base version (will not "forget" important pre-trained knowledge), and disadvantages, as it is able to adapt less to the given task (Biderman, Dan, et al., 2024).

We split out the dataset into three pieces: *train*, *eval*, *and test* with a 95:5:5 ratio respectively. The first is used for training our model, the second is used for in-training loss evaluation, and the latter is used in the test section, where we showcase the final loss of our model.

We showcase the prompt format of the LLaMA model in Table 5. The model is trained to complete the text in continuation of the "assistant" tag. Hyper-parameters are also available in Table 5. Note that we stop training after 1 epoch as loss from measured on the evaluation split converges rather quickly, and ultimately plateaus (Figure 1). Also note that we dispense samples with a length of over 2200 tokens (measured with the model's BPE tokenizer), because of the limited GPU memory we have available.

| | |
|---|---|
| System | `<|start_header_id|>system<|end_header_id|>`<br><br>`Ești un asistent util care analizează texte și autorii lor.<|eot_id|>` |
| User | `<|start_header_id|>user<|end_header_id|>Generează cuvinte-cheie din următorul text, care să aibă legătură atât cu conținutul textului, cât și cu trăsături ale autorului.`<br><br>`Puteți,va rog,sa mi explicați teorema cosinusului pe scurt,sunt în clasa a 7a și vreau sa învăț cât mai repede și eficient`<br><br>`<|eot_id|>` |
| Model | `<|start_header_id|>assistant<|end_header_id|>fizica, stiinta, interes stiintific` |

| Hyperparameter value | Value |
|---|---|
| Effective Batch Size | 8 |
| Number of epoch | 1 |
| Learning rate | 0.0002 |
| Warmup steps | 10 |
| Weight decay | 0.0 |
| Maximum sequence length | 2200 |
| LoRA rank | 16 |
| LoRA alpha | 16 |
| LoRA dropout | 0.05 |

**Table 5.** Prompt format example for Llama 3.1-8B-Instruct (actual text and keywords may vary), and hyperparameters used for training

### 4.2 Model Testing

On the *test* split, we employ two ways of measuring our model's success: exact match and Intersection over Union (IoU).

First, for every evaluated sample, we compare the set of generated keywords by our model, and the set of expected keywords (according to the corpus). We obtain an accuracy of 78.5%.

Second, we acknowledge that our corpus annotation might be prone to errors, in the sense that some keywords may not be evidently related to the text just based on the subreddit section, and thus are very hard for our model to infer. Consequently, we apply a more forgiving metric, measuring the ratio between the cardinality of the intersection and the union of the two sets of keywords (predicted and expected). On average, on the *test* split, we obtain an 80.4% IoU average.

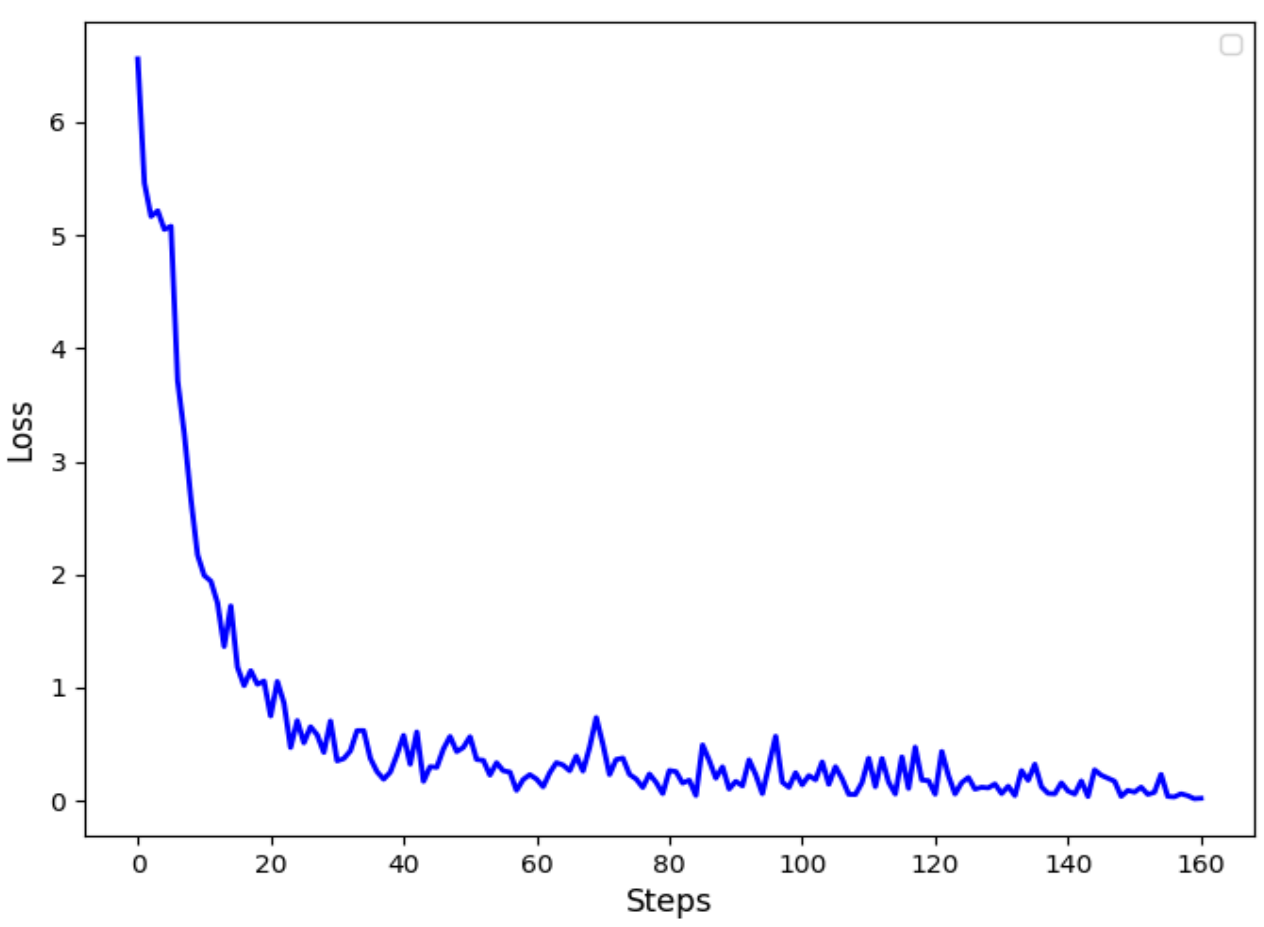

**Figure 1.** Loss over time while training, measured on the evaluation split of the dataset

### 4.3 Other Model Statistics and Behaviours

We observe the model's tendency to generate slightly fewer keywords per sample. The reference samples yield a 2.31 average of keywords, whilst the generated outputs yield a 2.26 average.

Moreover, there are 53 unique expected keywords in the *test* split. Putting together all of our model's predictions, we obtain 52 unique generated keywords; all of them are included in the expected corpus keywords, proving that the model is not yet able to generate new types of information (that would have not been encountered in the fine-tuning phase, but would have rather been inferred from its pre-training phase) about an author.

Based on this paper's authors' manual review, we make the observation that the model tends to generate slightly more content-focused keywords, rather than authored-focus. This pattern appears in our dataset as well but to a lesser degree. We are unable to quantify this phenomenon, but we note that the difference is not major, as authorship profiling remains the model's main focus.

Lastly, we observe the model's inability to combine various scenarios from the training phase. For example, in our dataset, most of the samples which flag the author as a speaker of a certain subdialect, do not have any other additional keywords included, due to the limitations of our annotation methodology. This behaviour is also observed in the model's generated keywords. Predictions related to the spoken subdialect are rarely accompanied by other keywords. For instance, the model generates "grai ardelenesc" (English: "Transilvanian subdialect") about the author of the following text: "Gulas bun in Budapesta. Intreb clujenii ca mno e o comunitate mai mare de cunoscatori" (English: "Good stew in Budapest. I ask the people of Cluj because there is a larger community of connoisseurs"). However, the keyword "food interest”, which is found in some other samples of our dataset, could have been suitable to be generated, but it is not.

All in all, the accuracy and IoU are promising for this baseline approach, and suggest how further research about Romanian authorship profiling has a lot of potential if further research is done.

## 5 Conclusion and Future Work

This paper introduced the first authorship profiling-oriented corpus for the Romanian language, with text samples annotated with keywords from various categories. Moreover, we train a baseline LLM for a keyword generation task, a problem that can also be viewed as authorship feature extraction. The promising, yet limited, capabilities of the model suggest how more research is needed in this field to obtain better results. In the future, we aim to obtain a more diverse dataset, that is not only sourced from Reddit. More robust methodologies for data annotation should be employed, as our approach is prone to some errors (limitations). Lastly, other model architectures, sizes, and fine-tuning methodologies should be experimented with.

## 6 Limitations and Ethical Concerns

First, the tags were manually extracted based on the characteristics of the subreddit. This is a highly biassed process, susceptible to personal assumptions and stereotypes. Although not documented, this could potentially generate unwanted toxic behaviours in our model. Moreover, the process of annotating almost all subreddit posts with the same tags is prone to error. Sometimes, certain tags (specific to a given subreddit) are impossible to infer by a model from the given text. More thoroughness, diversity, and manual evaluation are needed to mitigate these issues.

Secondly, although our classification is diverse (tackling demographic traits and personality characteristics at the same time), it comes with its downsides. Keywords much more closely related to the characteristics of the author, such as the status, get intertwined with keywords targeting just the text’s content. Oftentimes, the line between the two can get blurry and hard to distinguish.

Lastly, our computing resources are limited, so we were not able to perform extensive experiments with other models, or even full fine-tuning.